\def\year{2021}\relax
\documentclass[letterpaper]{article} 
\usepackage{aaai21}  
\usepackage{times}  
\usepackage{helvet} 
\usepackage{courier}  
\usepackage[hyphens]{url}  
\usepackage{graphicx} 
\usepackage{natbib}  
\usepackage{caption} 
\usepackage{amsthm}
\usepackage{amsmath}
\usepackage{multirow}
\usepackage{subfigure}
\usepackage[dvipsnames]{xcolor}
\usepackage[ruled,vlined]{algorithm2e}
\usepackage[switch]{lineno}
\title{CommPOOL: An Interpretable Graph Pooling Framework for Hierarchical Graph  Representation Learning}
\author {Haoteng~Tang,\textsuperscript{\rm 1}
        Guixiang~Ma, \textsuperscript{\rm 2}
        Lifang~He \textsuperscript{\rm 3} 
        Heng~Huang \textsuperscript{\rm 1} 
        Liang~Zhan \textsuperscript{\rm 1} \\
}
\affiliations {
    \textsuperscript{\rm 1} Department of Electrical and Computer Engineering, University of Pittsburgh, Pittsburgh, USA\\
    \textsuperscript{\rm 2} Intel Labs, Hillsboro, USA \\
    \textsuperscript{\rm 3} Department of Computer Science and Engineering, Lehigh University, Bethlehem, USA  \\
    hat64@pitt.edu, guixiang.ma@intel.com, lih319@lehigh.edu, heng.huang@pitt.edu, liang.zhan@pitt.edu
}

\begin{document}

\maketitle
\begin{abstract}
Recent years have witnessed the emergence and flourishing of hierarchical graph pooling neural networks (HGPNNs) which are effective graph representation learning approaches for graph level tasks such as graph classification. However, current HGPNNs do not take full advantage of the graph's intrinsic structures (e.g., community structure). Moreover, the pooling operations in existing HGPNNs are difficult to be interpreted. In this paper, we propose a new interpretable graph pooling framework - CommPOOL, that can capture and preserve the hierarchical community structure of graphs in the graph representation learning process. Specifically, the proposed community pooling mechanism in CommPOOL utilizes an unsupervised approach for capturing the inherent community structure of graphs in an interpretable manner. CommPOOL is a general and flexible framework for hierarchical graph representation learning  that can further facilitate various graph-level tasks. 
 Evaluations on five public benchmark datasets and one synthetic dataset demonstrate the superior performance of CommPOOL in graph representation learning for graph classification compared to the state-of-the-art baseline methods, and its effectiveness in capturing and preserving the community structure of graphs.         

\end{abstract}
\section{Introduction}
\noindent In recent years, 
Graph Neural Network (GNN) has emerged and been broadly used as a generalized deep learning architecture for graph representation learning in many fields, such as social network analysis \cite{chen2018fastgcn,huang2018adaptive}, chemical molecule studies \cite{dai2016discriminative,duvenaud2015convolutional,gilmer2017neural} and brain network analysis \cite{ma2019deep,liu2019community}. Generally, GNN models learn node embeddings by passing, transforming and aggregating node features across the graph. The generated node representations can then be forwarded to further layers for specific learning tasks, such as node classification \cite{kipf2016semi,velivckovic2017graph} and link prediction \cite{kipf2016variational}.. 

\begin{figure}[h]
\vspace{-0.4em}
    \vspace{-0.0001in}
    \centering
    \raggedright
    \begin{minipage}[h]{0.6\linewidth}
    \centering
    \includegraphics[width=3.2in]{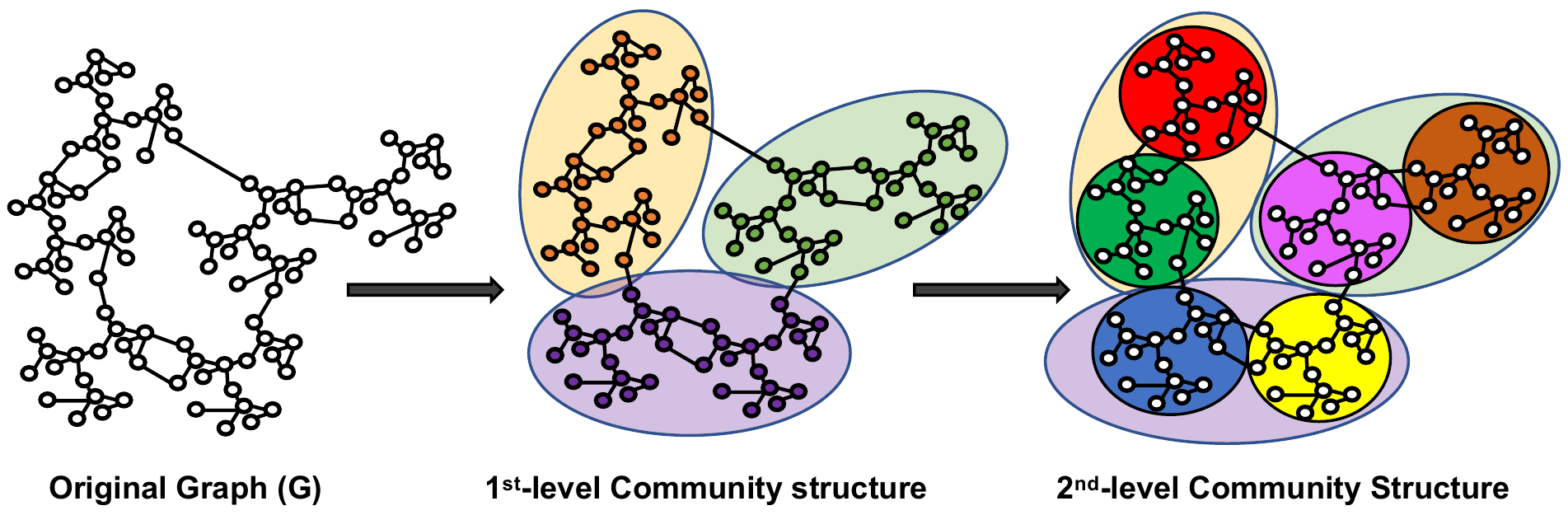}
    \end{minipage}
    \caption{An example of hierarchical community structures in graphs}
    \label{Hierarchical}
    \vspace{-0.0001in}
    \vspace{-0.4em}
\end{figure}

Most of the existing GNN models (e.g., GCN \cite{kipf2016semi}, GAT \cite{velivckovic2017graph}, GraphSage \cite{hamilton2017inductive}) focus on node-level representation learning and only propagate information across edges of the graph in a flat way. When applying these GNNs for graph-level tasks such as graph classifications, existing works usually apply simple global pooling strategies (i.e., a summation over the learned node representations) to obtain the graph-level embedding and use it for graph label prediction  \cite{li2015gated,vinyals2015order,zhang2018end} or graph similarity learning \cite{ma2019survey}. One main drawback in these GNNs is that the hierarchical structure, often existing in graphs, is ignored during the global pooling process, which makes the models less effective for graph-level tasks. Hierarchical structure is a very important structure for many graphs in various domains. For example, the hierarchical community structure shown in \textbf{Figure~\ref{Hierarchical}} is a typical pattern that often appears in social networks \cite{girvan2002community,long2019hierarchical}, chemical molecule networks \cite{spirin2003protein} and brain networks \cite{kong2014brain,meunier2009hierarchical}. Therefore, preserving these community structures is critical for better understanding and analyzing these graphs.

Some recent works proposed hierarchical graph pooling neural networks (HGPNNs) to address the hierarchical structure representation issue by introducing the hierarchical pooling operations \cite{ying2018hierarchical,lee2019high,gao2019graph,zhang2019hierarchical}. Generally, these HGPNNs consist of two components: the GNN backbone which is used to embed the graph nodes and local structures, and the pooling operation which represents graph structure in a hierarchical way. 
These HGPNNs have demonstrated the necessity of adding hierarchical pooling operations in GNNs to better preserve the graph hierarchical structure. 

However, a critical limitation of the existing hierarchical graph pooling (HGP) strategies is that few of the pooling operations in the models are interpretable. In many real applications, it is desirable to have an interpretable model, where human can understand the cause of a decision made by the model \cite{miller2019explanation,molnar2020interpretable}. Moreover, an interpretable model is more robust under adversarial attacks \cite{dai2018adversarial,zugner2018adversarial,zugner2020certifiable,tang2020adversarial}. 
A few of recent works \cite{ying2019gnnexplainer,hou2019measuring,yuan2020xgnn} interpreted the node feature embedding via GNN as a neighborhood aggregation scheme. 
Particularly, they stated that the GNN embed the local feature of each node $v$ within two steps: (1) neighbor node features aggregation and (2) node feature transformation. 
However, the interpretability of pooling operations is still not well solved.
(Details are discussed in the \textbf{Related Work}). In order to make the HGP operation interpretable, three questions should be considered: \\
\textbf{Q1}: How to capture the graph hierarchical structures in an interpretable way?\\ 
\textbf{Q2}: How to scale down the graph representation while preserving the structures via an interpretable process? \\
\textbf{Q3}: What do we obtain after the pooling operation? 

To address these challenges, we propose a Community-Based HGP framework, \textbf{COMMUNITY-POOL} or \textbf{CommPOOL}. We aim to encode the hierarchical community structure in graphs, which is a natural structure in many graphs, where nodes within each community are more densely connected than the nodes across different communities. Specifically, we propose a community-based hierarchical pooling operation
which aggregates and synthesizes the node features based on the detected communities, such that the community structure of graphs can be preserved during the pooling process. Moreover, we introduce a GNN-based framework with the proposed community-based hierarchical pooling operation for learning latent graph representations, where both local node features and the hierarchical community-structure information are encoded and preserved. 
Our contributions here can be summarized as:
\begin{itemize}
\item We propose a community-based HGP framework (CommPOOL) for learning graph representation in a hierarchical way that can preserve both the local node features and the hierarchical community structure of graphs.
\item The proposed hierarchical community pooling strategy relies on the community structure which is explicitly detected from the graphs, therefore the pooling operation can capture the intrinsic community-level latent representation of graphs and the pooling process is inherently interpretable. 
\item We evaluate our CommPOOL framework for the whole graph classification task on multiple public benchmark datasets. The results demonstrate the superior performance of our model compared to several state-of-the-art graph pooling neural networks. 
\item Evaluations on synthetic graphs with community ground-truth labels show that our proposed CommPOOL can capture and preserve the intrinsic community structure of graphs during the learning process.
\end{itemize}

\section{Related Work}
\subsection{Graph Pooling}
Graph pooling operation is a strategy aiming to scale down the size of input graphs. It can not only help to avoid model overfitting and reduce the computational cost but also generate graph-level representations \cite{wu2020comprehensive}. In the early works \cite{henaff2015deep,levie2018cayleynets,dhillon2007weighted,vinyals2015order}, 
the graph pooling methods simply compute the mean/max/sum of all graph node features as the representation of the whole graph. Such a primitive pooling strategy is named as global pooling. Later on, a few advanced techniques (e.g. attention mechanisms \cite{li2015gated,gilmer2017neural,tran2018filter}, feature sorted \cite{zhang2018end}) are proposed to improve the performance and efficiency of the global pooling.  
However, the global pooling methods do not learn the hierarchical representations, which are crucial for capturing the structural information of graphs. Therefore, HGPNNs are proposed. 

 
\subsection{Interpretability of HGPNNs}
Most HGP operations in the current HGPNNs \cite{zhang2019hierarchical,ying2018hierarchical,lee2019self,gao2019graph,kefato2020graph,Bianchi2020SpectralCW} show little interpretability and are difficult to be understood by the users. Moreover, very few studies present the interpretabiliy of their HGP operations in the paper, which may be accounted for by the following two issues: 
\textbf{(1).} Hardly any clear definition or analysis can be found to explain what is the captured graph structure. Therefore, the model users are lack of heuristic knowledge to understand the pooling operation. \textbf{(2).} Although some studies \cite{ying2018hierarchical} present the visualization of hierarchical clusters captured by the model, no quantitative analysis is provided to examine whether the captured clusters of nodes are aligned with the intrinsic clusters in the original graph. Apart from these, most recent studies unfold the HGP operation as a neural network layer with trainable parameters. The black-box nature of neural networks may also raise extra difficulties to interpret the models in a way. 
\section{Preliminaries}
\subsection{Graph Notation} 
We consider the graph classification problem on attributed graphs with different numbers of nodes. Let $G=(A, H)$ be any of the attributed graph with $N$ nodes, where $A\in\{0,1\}^{N \times N}$ is the graph adjacency matrix and $H \in \mathcal{R}^{N \times d}$ is the node feature matrix assuming that each node has $d$ features. Also, $Z=[Z_{1},...,Z_{N}]^{T}$ is defined as the node latent feature matrix where $Z_{i}$ is the latent feature vector for the node $i$. Given a set of labeled data $\mathcal{D} = \{(G_{1},y_{1}), (G_{2},y_{2}), (G_{3},y_{3}), ...\}$  where $y_{i} \in \mathcal{Y}$ is the classification label to the corresponding graph $G_{i} \in \mathcal{G}$. The graph classification task can be formulated as learning a mapping, $f$: $\mathcal{G} \rightarrow \mathcal{Y}$.

\subsection{Graph Neural Network}
Graph Neural Network (GNN) is an effective message-passing architecture for embedding the graph nodes and their local structures. Generally, GNN can be formulated as:
\begin{equation}
    Z^{(k)} = F(A^{(k-1)}, Z^{(k-1)}; \theta^{(k)}),
\end{equation}
where $k$ denotes the layer $k$ of GNN. 
$A^{(k-1)}$ is the graph adjacency matrix computed by layer $(k-1)$ of the GNN. $\theta^{(k)}$ is the trainable parameters in the layer $k$ of the GNN. Particularly, $Z^{0}=H$.

$F(\cdot)$ is the forward function to combine and transform the messages across the nodes. Many different versions of forward functions $F(\cdot)$ are proposed in the previous studies \cite{gilmer2017neural,hamilton2017inductive} such as Graph Convolutional Neural Network (GCN) \cite{kipf2016semi} and Graph Attention Network (GAT) \cite{velivckovic2017graph}. The GCN linearly combines the neighborhoods as the node the representation. And the GAT computes node representations in entire neighborhoods based on attention mechanisms \cite{bruna2013spectral}.

\begin{figure*}[h]
\small
    \vspace{-0.0001in}
    \centering
    \raggedright
    \begin{minipage}[h]{0.6\linewidth}
    \centering
    \includegraphics[width=7in]{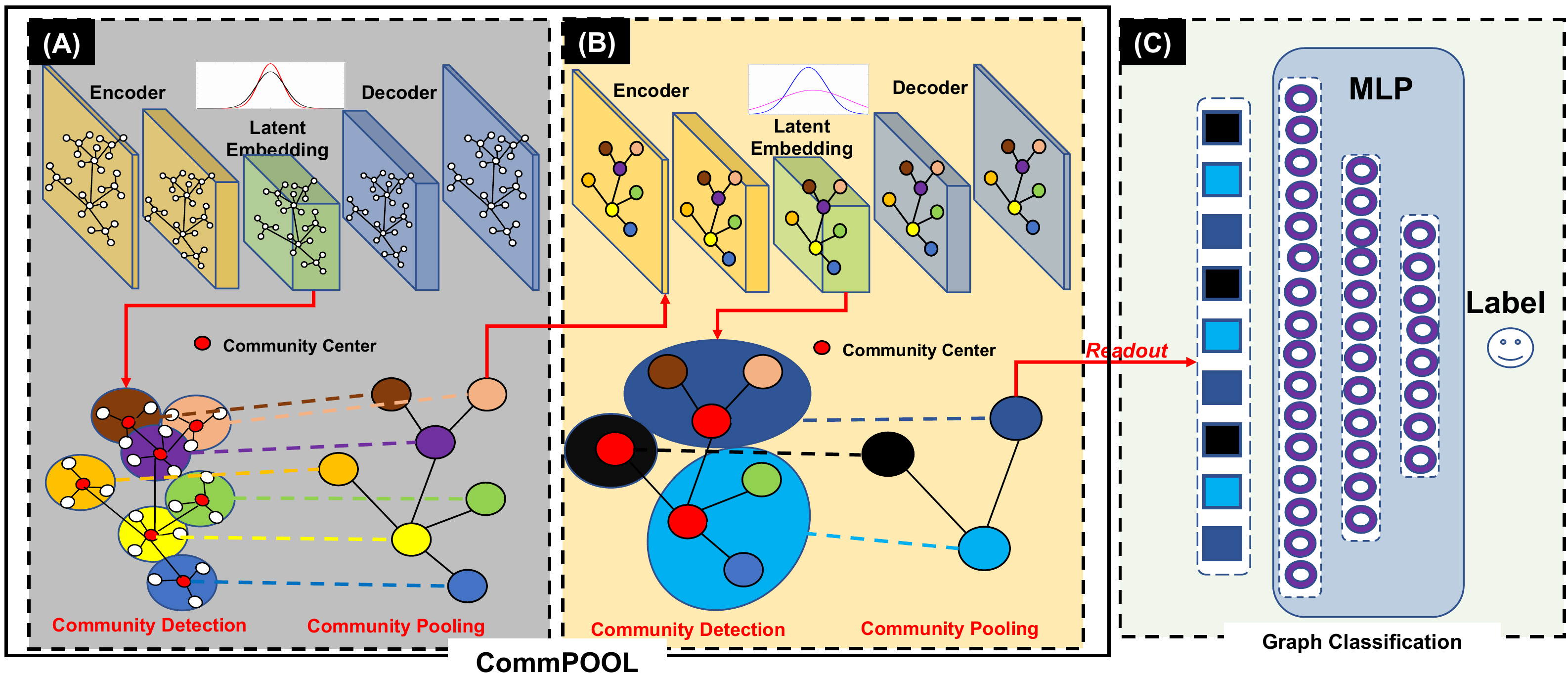}
    \end{minipage}
    \caption{Framework of the CommPOOL for graph classification. \textbf{(A)} is the $1^{st}$ Embedding-Pooling (EP) module and \textbf{(B)} is the $2^{nd}$ EP module. In each module, we embed the graph into the latent space by using VGAE. In the latent space, we scale down the graph representation based on the detected communities. \textbf{(C)} is the MLP for graph classification.}
    \label{Architecture}
    \vspace{-0.0001in}
\end{figure*}
\section{The Proposed Framework}
\subsection{Model Architecture}
Our goal is to provide a general graph pooling framework that can    capture and preserve the hierarchical community structure of graphs in the representation learning process of GNNs. The framework should be interpretable and it should be able to facilitate further graph-level learning tasks, for example, graph classification. To achieve this goal, we propose a community-based hierarchical graph pooling (HGP) framework: \textbf{CommPOOL}, which is composed of $k$ cascaded Embedding-Pooling (EP) modules to learn the graph representation in a hierarchical way. Each EP module consists of (1) an Embedding stage, where a GNN model is employed to get the latent node representations (i.e., node embeddings) of the input graph, and (2) a Pooling stage, where a newly proposed community pooling mechanism is conducted on the node embeddings to detect communities from the graph and obtain a scaled-down graph-level representation that encodes both the local node features and the community structure of the graph. The output of the last EP module will be the final graph-level representation that preserves the overall hierarchical community structure of the graph. \textbf{Figure~\ref{Architecture}(A, B)} shows an instance of the proposed framework with two cascaded EP modules. In real applications of our framework, the choice of value for $k$ is flexible and it can be decided based on the practical needs or domain knowledge for the specific application (e.g., domain evidence about how many community hierarchies exist in the graphs). In this paper, we set $k = 2$ and use the architecture given in \textbf{Figure~\ref{Architecture}(A, B)} for illustrating our framework and we use the MLP shown in \textbf{Figure~\ref{Architecture}(C)} for evaluating the CommPOOL in graph representation learning for facilitating graph classification task.   

In the following subsections, we introduce the two main parts in the proposed EP module for CommPOOL: (1) the GNN-based Graph Node \textbf{Embedding}, and (2) the Community \textbf{Pooling} Operation. 
\subsection{GNN-based Node Embedding}
We aim at a general GNN-based model to embed the graph nodes into the latent feature space $\mathcal{Z}$ that well preserves the inherent graph structures. On the one hand, the desired node latent features should well encode the node information and the information between the node and its neighbors. On the other hand, the latent features should preserve the intrinsic structures of the graphs without task-specific influences or supervised information. 
On account of the above considerations, we choose the Variational Graph Auto-Encoders (VGAE) \cite{kipf2016variational} to embed nodes into the latent space by reconstructing the graph itself.   
\subsubsection{Encoder}
In the VGAE, we need to learn a Gaussian distribution $q(Z|H,A) = \mathcal{N}(Z|\mu,\sigma^{2})$ which is used to approximate the Gaussian prior $p(Z)=\mathcal{N}(Z|0,I)$. Particularly, we utilize two GNN layers to compute the $\mu$ and $\sigma^{2}$ parameters of $q$. In the first layer, $\mu$ and $\sigma^{2}$ share the same GNN encoder. And in the second layer, two separate GNNs are used to generate $\mu$ and $\sigma^{2}$ respectively. The approximation can be achieved by maximizing the Kullback–Leibler ($KL$) loss between $p$ and $q$:
\begin{eqnarray}
\mathcal{L}_{KL} = KL(q(Z|H,A)||p(Z))
\end{eqnarray}
The latent features $Z$ can be obtained by resampling from the optimal $q(Z|H,A)$.  
\subsubsection{Decoder}
After we obtain the latent features $Z$, we reconstruct the original graph adjacency matrix by:
\begin{eqnarray}
\hat{A} = sigmoid(ZZ^{T}).
\end{eqnarray}
We define $+$ and $-$ as the edges and non-edges position index in $A$. So we reconstruct  the adjacency matrix by minimizing the $\mathcal{L}_{A}$:
\begin{eqnarray}
\mathcal{L}_{A} &=& \mathcal{L}^{+} + \mathcal{L}^{-} \\ 
            &=& -\frac{1}{E_{1}} \Sigma(log(\hat{A}^{+})) - \frac{1}{E_{2}} \Sigma (log(1-\hat{A}^{-})) \nonumber 
\end{eqnarray}
where $E_{1},E_{2}$ is the number of edges and non-edges. The overall objective function of VGAE is:
\begin{eqnarray}
\underset{Z \in \mathcal{Z}}{\text{minimize}} \quad \mathcal{L}_{A} - \mathcal{L}_{KL}
\end{eqnarray}
In our CommPOOL, we use GCN \cite{kipf2016semi} to build up the basic encoder layers and use GAT \cite{velivckovic2017graph} as the encoder variations.
\subsection{Community Pooling}
\subsubsection{Community Capturing}
After embedding the graph nodes into the latent space $\mathcal{Z}$, we adopt an unsupervised clustering method Partitioning Around Medoids (PAM) \cite{kaufmann1987clustering,lee2019high} on the node latent feature vectors to group the graph nodes into $L$ different communities, where $L$ is a parameter denoting the number of communities in the graph.
Our community partition problem can be defined as: given all the $N$ nodes in graph $G$ with their latent feature vectors set $V=\{Z_{1},..., Z_{N}\}$, find $L$ different nodes with their latent features $Z_{C}=\{Z_{C_{1}},..., Z_{C_{L}}\} \subset V$ from the $N$ nodes as the optimal community centers, and assign the other nodes into these $L$ communities based on the distances between their latent feature vectors ($O = V \setminus Z_{C}$ ) and $Z_{C}$. PAM realize the community partition problem via the following four steps. 
\begin{itemize}
    \item \textbf{Step 1.} Initialization: Randomly select $L$ nodes with their features $Z_{C}$ as the community medoid nodes.
    \item \textbf{Step 2.} Clustering: Compute the $L_{1}$ distances between the medoid nodes and the rest nodes based on their feature vectors, and assign each non-medoid node to its closest community; Calculate the value for the below cost function, which computes the total distance between the non-medoid node feature vectors $O_{j}\in O$ and their community medoid feature vectors by: 
    \begin{eqnarray}
    Cost = \Sigma_{j=1}^{N-L}|O_{j}-Z_{C_{x}}|_{L_{1}}, 
    \end{eqnarray}
    where $Z_{C_{x}}\in Z_{C}$ is the corresponding medoid of $O_{j}$. 
    \item \textbf{Step 3.} Adjusting: Swap each medoid node by all other non-medoids  
    and calculate the total cost for current configuration referring to Step 2. Compare the cost of current and previous configuration and keep the configuration with the smaller total cost. 
    \item \textbf{Step 4.} Optimization: Repeat Step 2 and 3 until the configuration does not change. 
\end{itemize}

\begin{table*}[h]
\centering
\caption{Average graph classification test accuracy $\pm$ standard deviation ($\%$).}
\begin{tabular}{c|c|c|c|c|c}
\hline
Models & BZR & Synthie & FRANKENSTEIN & PROTEINS & AIDS\\
\hline
Set2Set & $80.50\pm1.03$ & $22.50\pm0.86$ & $60.62\pm0.27$ & $68.08\pm0.56$ & $88.80\pm0.45$ \\
SortPool & $77.00\pm1.24$ & $32.50\pm1.24$ & $59.86\pm1.22$ & $70.11\pm0.04$ & $86.00\pm2.42$ \\
\hline
DIFFPOOL & $80.50\pm 1.48$ & $57.00\pm 2.62$ & $60.60\pm 1.62$ & $72.43\pm 0.26$ & $93.50\pm 1.00$ \\
SAG-POOL & $82.00\pm 2.13$ & $45.00\pm 4.21$ & $61.73\pm 0.76$ & $71.86\pm 0.97$ & $93.50 \pm 1.00$ \\
HGP-SL & $83.00\pm 4.30$ & $54.00\pm 0.04$ & $59.51\pm 1.50$ & $\textbf{84.91}\pm \textbf{1.62}$ & $95.50\pm 1.00$ \\
\hline
CommPOOL & $\textbf{86.00}$ $\pm$ $\textbf{1.23}$ & $\textbf{66.50}$ $\pm$ $\textbf{0.38}$ & $\textbf{62.15}$ $\pm$ $\textbf{0.37}$ & $74.74$ $\pm$ $0.06$ & $\textbf{98.50}$ $\pm$ $\textbf{0.05}$ \\
\hline
\end{tabular}
\label{accuracy}
\end{table*}

\subsubsection{Community Pooling}
In order to preserve the captured community structure  during the pooling process for the entire-graph representation learning, we propose a new pooling mechanism called ``community pooling", which summarizes the learned node representations based on the detected community structure. Suppose $Z_{M_{i}} = \{Z_{M_{i}}^{1}, ..., Z_{M_{i}}^{W}\}$ is the set consisting of the latent feature vectors of all $W$ community member nodes except for the community center nodes $Z_{C_{i}} \in Z_{C}$ in the community$-i$. Our community pooling problem can be defined as: given a community center feature $Z_{C_{i}} \in Z_{C}$, and the corresponding $W$ community member features $Z_{M_{i}}$, compute the community representation $Z_{Comm_{i}}$. The community pooling operation computes the community$-i$'s representation by:
\begin{eqnarray}
Z_{Comm_{i}} = Z_{C_{i}} + \Sigma_{w=1}^{W} \ Sim(Z_{M_{i}}^{w},Z_{C_{i}}) Z_{M_{i}}^{w},
\end{eqnarray}
where $Sim(\cdot)$ is a function to measure the similarity between each member $Z_{M_{i}}^{w}$ and the community center $Z_{C_{i}}$. In our model, we mainly define $Sim(\cdot)$ based on $L_{1}$ distance:
\begin{eqnarray}
Sim(Z_{M_{i}}^{w},Z_{C_{i}}) = \frac{1}{\| Z_{M_{i}}^{w}-Z_{C_{i}} \|_{L1}}
\end{eqnarray}
When each community representation $Z_{Comm_{i}}$ is computed, we replace the center node feature $Z_{C{i}}$ by $Z_{Comm_{i}}$ and remove other community member nodes. As for the graph topology structure, the preserved center nodes are connected if and only if they are connected in the original graph. To sum up, during the pooling, the community structure information and the node features are preserved onto the community center nodes. And the graph structures among the communities are presented as the topology structure of down-scaled graph with $M<N$ nodes. 
\subsection{CommPOOL for Graph Classification}
When the community representations $Z^{(K)}_{Comm}=[Z^{(K)}_{Comm_{1}},...,Z^{(K)}_{Comm_{L}}]^{T}$ are obtained from the last Embedding-Pooling module ($k=K$), a global readout operation is used to generate the whole graph representation $Z_{graph}$ by averaging $Z^{(K)}_{Comm}$. Finally, an Multilayer Perceptron (MLP) utilizes $Z_{graph}$ to make predictions for graph classification. The training procedure of CommPOOL for the graph classification task is summarized in \textbf{Algorithm~\ref{Train Procedure}}. 
\vspace{-0.5em}
\begin{algorithm}[h]
\small
\SetAlgoLined
\SetKwData{U^{*}}{U^{*}}\SetKwData{This}{this}\SetKwData{Up}{up}
\SetKwFunction{Union}{Union}\SetKwFunction{FindCompress}{FindCompress}
\SetKwInOut{Input}{Input}
\SetKwInOut{Output}{Output}
\Input{graph: $G=(A,H)$, classification label: $y$, $K$}
\Output{prediction: $\hat{y}$}
\For{$k=1,2,...,K$}{
\textbf{Step 1:} Use $G$ to train the VGAE \\
\textbf{Step 2:} Obtain the latent feature using trained VGAE \\
\textbf{Step 3:} Community Pooling on latent features and generate down-scaled graph $G^{(k)}=(A^{(k)},Z_{Comm}^{(k)})$.
Set $G=G^{(k)}$.} 
\textbf{Step 4:} $Z_{graph} = Global Readout (Z_{Comm}^{(K)})$ \\
\textbf{Step 5:} Train $MLP$ to generate $\hat{y} = MLP(Z_{graph})$ 
\caption{Training Procedure}
{\label{Train Procedure}}
\end{algorithm}
\vspace{-0.5em}
\section{Experiment}
In this section, we evaluate our CommPOOL framework using graph classification tasks. We present our experiment results in the following four subsections: \textbf{(1)} We introduce the dataset used in the experiments. \textbf{(2)} We compare the graph classification performance between CommPOOL and several competing HGPNN models. 
\textbf{(3)} We provide some variations of the CommPOOL. \textbf{(4)} We test our model on the simulation data to evaluate whether CommPOOL can accurately preserve the community structures in the graph.

\paragraph{\textit{Dataset.}}
Five graph dataset are selected from the public benchmark graph data collection \cite{KKMMN2016}. \textbf{Table~\ref{dataset}} summarizes the statistics of all dataset.

\textbf{PROTEINS} and \textbf{Synthie} \cite{borgwardt2005protein,dobson2003distinguishing,morris2016faster} are two sets of graphs representing the protein structure. The nodes are some amino acid features such as secondary structure content and amino acid propensities. Nodes are linked by edges if the amino acid is an amino acid sequence. \textbf{FRANKENSTEIN} \cite{orsini2015graph} is a set of graphs representing the molecules with or without mutagenicity. The nodes represent different chemical atoms and the edges are the chemical bonds type. \textbf{BZR} \cite{sutherland2003spline} is a set of graphs representing the ligands for the benzodiazepine receptor. And \textbf{AIDS} \cite{riesen2008iam} is set of graphs representing molecular compounds with activity against HIV or not. The molecules are converted into graphs by representing chemical atoms as nodes and the bonds as edges.



\begin{table}[h]
\small
\centering
  \caption{Dataset Statistics: $\mathcal{V}$ and the $\mathcal{E}$ represent the nodes and edges in graph $G$. $c$ represents graph classes.}
  \setlength{\tabcolsep}{1.5mm}{
  \label{table 1}
  \begin{tabular}{c|c|c|c|c}
  \hline
    Dataset & $\#|G|$ & Ave.$|\mathcal{V}|$ & Ave.$|\mathcal{E}|$ & $\#|c|$ \\
  \hline
    BZR & 405 & 35.75 & 38.36 & 2 \\
    Synthie &400 & 95.00 & 172.93 & 4 \\
    FRANKENSTEIN & 4337& 16.90 & 17.88 & 2 \\
    PROTEINS & 1113 & 39.06	& 72.82 & 2 \\
    AIDS & 2000 & 15.69 & 16.20 & 2 \\
  \hline
\end{tabular}}
\label{dataset}
\end{table}

\subsection{Graph Classification}
\subsubsection{Baseline Methods}
Our baseline methods include: two graph global pooling models (\textbf{Set2Set} \cite{vinyals2015order} and \textbf{SortPool} \cite{zhang2018end} ), and three HGP models (\textbf{DIFFPOOL} \cite{ying2018hierarchical}, \textbf{SAGPOOL} \cite{lee2019high} and \textbf{HGP-SL} \cite{zhang2019hierarchical}). For fair comparisons, we set two embedding-pooling modules for all HGP models including three baseline HGPs and our CommPOOL. For the baselines, we follow the original hyperparameter search strategies provided in the related papers. 

\subsubsection{Experiments Setting}
Following previous works \cite{ma2019graph,ying2018hierarchical,zhang2019hierarchical}, we randomly split the whole dataset into training ($80\%$) set, validation ($10\%$) set and testing ($10\%$) set. We repeat this randomly splitting process 10 times, and the average test performance with standard derivation is reported in \textbf{Table~\ref{accuracy}}. 
We optimize the model via Pytorch Adam optimizer. For the VGAE in the first module, the learning rate (lr) and the weight decay (wd) are searched in $\{ 0.0001, 0.001, 0.005, 0.01, 0.05, 0.1\}$. The dimension of two latent GNN layers are $32$ and $16$. For the VGAE in the second module, the lr and wd are searched in $\{ 0.0001, 0.001, 0.005, 0.01\}$ and the dimension of two latent GNN layers are $64$ and $32$. In the community pooling operation, the number of communities is searched in $\{40\%, 50\%, 60\% \}$ of the number of graph nodes ($N$). The MLP consists of two fully connected layers with $64$ and $32$ neurons and a softmax output layer. The lr for training the MLP is searched in $\{0.001, 0.005, 0.01\}$. We stop training if the validation loss does not decrease for 50 epochs. Our experiments are deployed on NVIDIA Tesla P100 GPUs. We implement all baselines and CommPOOL using PyTorch \cite{paszke2017automatic} and the torch geometric library \cite{fey2019fast}.
\subsubsection{Summary of Results}
\textbf{Table~\ref{accuracy}} summarizes the classification performances of six models on five public datasets. Our CommPOOL outperforms all baselines in the graph classification task on almost all datasets, especially on the four-class data \textbf{Synthie}. For example, our CommPOOL shows about $5.11\%$ improvement in the classification accuracy comparing to all baselines on \textbf{BZR} data. This superiority of CommPOOL may be credited to its advanced mechanism for capturing and preserving the community structure in the pooling operation. Also, these results indicate that the community is a crucial hierarchical structure for learning the whole graph representation.




Moreover, \textbf{Table~\ref{accuracy}} shows that hierarchical pooling methods generally perform better than global pooling methods, which verifies that the hierarchical pooling can better capture the graph global representations. Among all baseline models, HGP-SL relatively performs better than others, which may be attributed to the structure learning (SL) operations in the model. On PROTEINS, HGP-SL performs the best among all baseline methods and even better than ours, which implies that the structural learning strategy in HGP-SL might be specifically suitable for PROTEINS data. 
\subsection{Model Variations}
To show the flexibility of CommPool, we compare several variations of CommPOOL on PROTEINS and FRANKENSTEIN data. As noted in \textbf{The Proposed Framework} section, GAT \cite{velivckovic2017graph} is used to replace GCN as a VGAE encoder variation. Moreover, instead of using the reciprocal of $L_{1}$ distance, we adopt the cosine-similarity as $Sim(\cdot)$ to measure the similarity between community members $Z_{M_{i}}$ and the corresponding community center $Z_{C_{i}}$ in community$-i$:
\begin{equation}
    Sim(Z^{w}_{M_{i}},Z_{C_{i}}) = \frac{Z^{w}_{M_{i}}Z_{C_{i}}}{\|Z^{w}_{M_{i}}\| \|Z_{C_{i}}\|}
\end{equation}
The performance of CommPool with different encoders and similarity measures are listed in \textbf{Table~\ref{Variations}}, which indicates that \textbf{GAT}, compared to \textbf{GCN}, has a better performance as the encoder in CommPOOL to embed the graph nodes. In addition, \textbf{Table~\ref{Variations}} shows that $L_{1}$ distance is better than $cosine$ distance when measuring the similarity between the latent features of community member nodes and the community center nodes. A possible explanation is that $L_{1}$ distance is used in the PAM clustering. Therefore, it may be better to use the same distance metric in the community partition process. 
\vspace{-0.5em}
\begin{table}[h]
\centering
  \caption{Performance (\%) of CommPOOL with different encoder settings and different similarity measures}
  \label{table 3}
  \setlength{\tabcolsep}{0.25mm}{
  \begin{tabular}{c|c|c|c}
  \hline
  \multicolumn{2}{c|}{\textbf{ CommPOOL }} & \textbf{  PROTEINS  } & \textbf{ FRANKENSTEIN } \\
  \hline
  \multirow{2}*{GCN}
  ~ & $L_{1}$ & $74.74\pm 0.06$ & $62.15\pm0.37$ \\
  ~ & $cosine$ & $73.84\pm0.13$ & $60.18\pm0.42$  \\
  \hline
  \multirow{2}*{GAT}
  ~ & $L_{1}$ & $78.84\pm0.02$ & $63.48\pm0.52$  \\
  ~ & $cosine$ & $76.01\pm0.21$ & $62.32\pm0.39$ \\
  \hline
\end{tabular}}
\label{Variations}
\end{table}
\vspace{-0.5em}
\subsection{Community Evaluation on Simulation Data}
Since no community ground-truth is provided in any publicly graph classification datasets, we simulate a set of graphs with the known community ground-truth and evaluate how CommPOOL preserves the intrinsic community structures on these simulation graphs.
\vspace{-0.5em}
\begin{table}[h]
    \centering
    \caption{Average graph classification accuracy $\pm$ standard deviation (\%) on the simulation data.}
    \label{tab:my_label}
    \begin{tabular}{c|c}
         \hline
         Models & Classification Accuracy \\
         \hline
         Set2Set & $46.54\pm3.85$ \\
         SortPOOL & $51.29\pm0.61$ \\
         \hline
         DIFF-POOL & $67.14\pm 2.16$ \\
         SAG-POOL &  N/A \\
         HGP-SL & $72.70\pm 1.95$ \\
         CommPOOL & $80.14\pm 2.15$ \\
         \hline
    \end{tabular}
    \label{simulation accuracy}
\end{table}
\vspace{-0.5em}
\paragraph{\textit{Simulation Graphs.}}
We create $3$ classes of simulation graphs using different graph generating methods, including the Random Partition Graphs, the Relax Caveman Graphs, and the Gaussian Random Community Graphs \cite{brandes2003experiments,fortunato2010community}. 
Each class contains $300$ graphs and each graph has $4$ communities with the average size of $6$ nodes. A community label is assigned to each graph node. Meanwhile, we randomly sample from the normal distribution $\mathcal{N}(0,I)$ as node features. We evaluate CommPOOL on the simulation graphs to predict their class labels. \textbf{Table~\ref{simulation accuracy}} compares the graph classification performance of CommPool with the baseline models. The results show that, on the simulation data, the CommPOOL can also outperform all the baseline models. \textbf{N/A} in \textbf{Table~\ref{simulation accuracy}} indicates the SAG-POOL cannot achieve an optimal point in reachable epochs. 

\begin{figure}[t]
\vspace{-0.5em}
    \subfigure[]{
    \centering
    \begin{minipage}[t]{0.9\linewidth}
    \centering
    \includegraphics[width=2in]{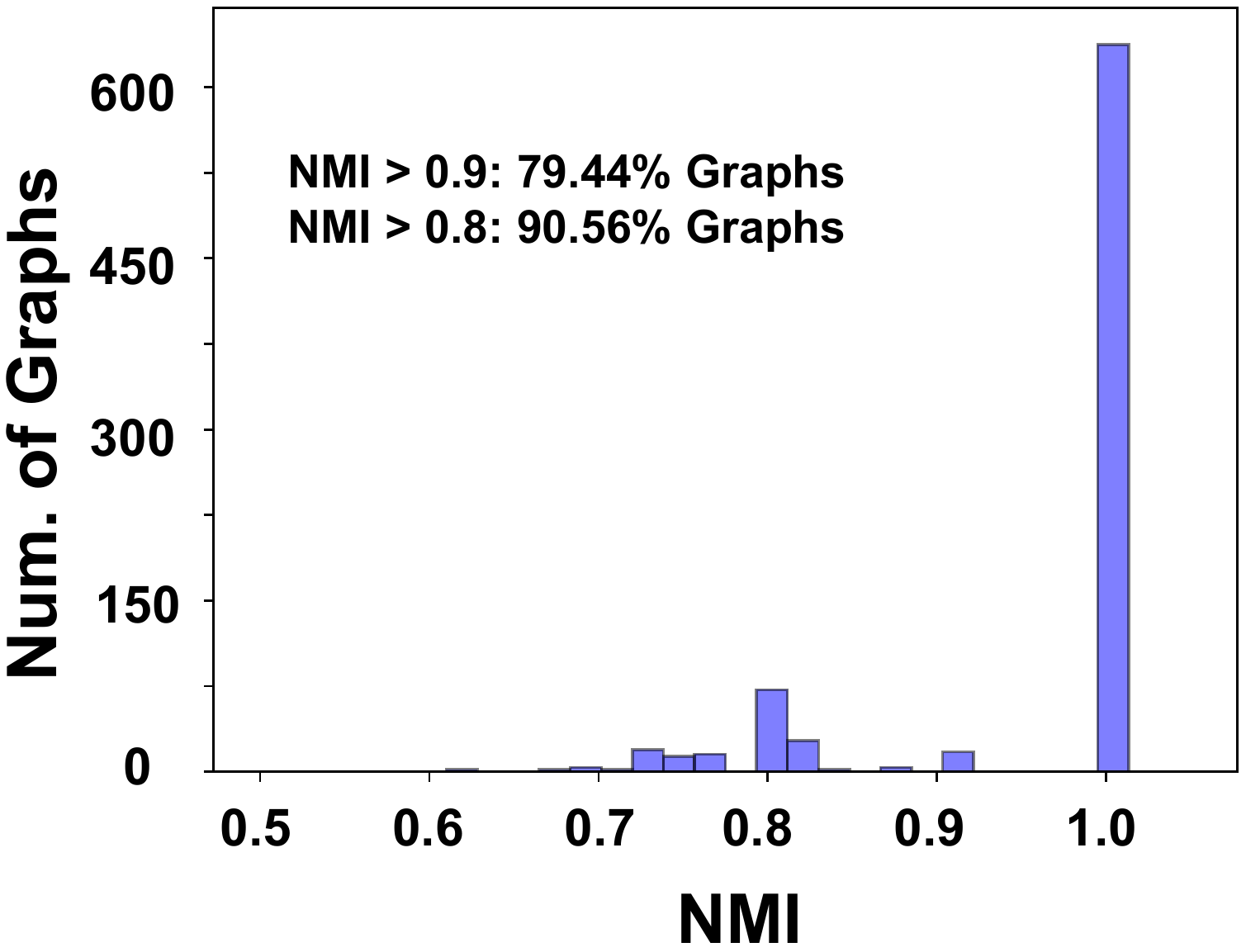}
    \end{minipage}
    }    
    \subfigure[]{
    \centering
    \begin{minipage}[t]{0.45\linewidth}
    \includegraphics[width=1.6in]{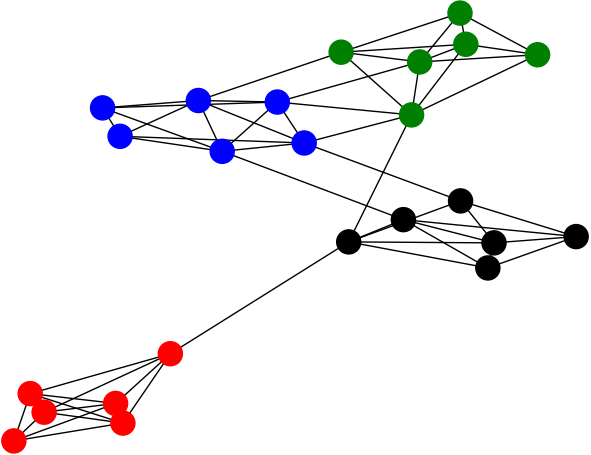}
    \end{minipage}
    }
    \subfigure[]{
    \centering
    \begin{minipage}[t]{0.45\linewidth}
    \includegraphics[width=1.6in]{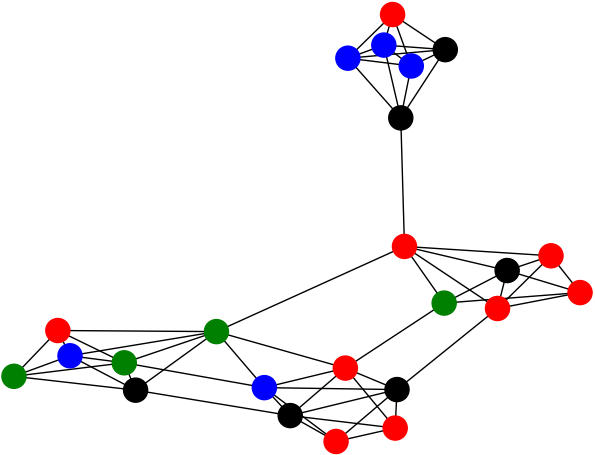}
    \end{minipage}
    }
    \caption{(\textbf{a}) is the NMI histogram between the distributions of community labels of ground-truth and CommPOOL prediction. \textbf{(b)} 
    is a positive example of community structure captured by CommPOOL leading to a correct classification. \textbf{(c)} is a negative example which leads to a graph misclassification. Different colors represent different communities.}
    \label{community}
\vspace{-0.5em}
\end{figure}

More importantly, in order to evaluate if CommPool can capture the community structures, we compare the node community label assigned by PAM clustering in the $1^{st}$ EP module to the community ground-truth labels. 
Specifically, we compute the \textbf{Normalized Mutual Information (NMI)} \cite{strehl2002cluster} between distribution of community labels predicted by the model and given by the ground-truth for each graph.  \textbf{Figure~\ref{community}a} is a histogram presenting the distribution of NMI scores for all $900$ simulation graphs. Statistically, $79.44\%$ graphs have an NMI score larger than $0.9$ and the mean NMI score is $0.952\pm0.098$. 
\section{Evaluation and Discussion}
In this section, we firstly discuss the interpretibility of our proposed community pooling operation. And then we analyze the importance of community structure to the graph classification task. 
\subsection{Interpretability of Community Pooling}
CommPOOL is a hierarchical graph pooling framework with an interpretable pooling operation. The user can transparently understand the pooling results by monitoring the pooling operation. An interpretable pooling operation should be capable of clearly answering three questions mentioned in the \textbf{Introduction} section. Our CommPOOL provides the heuristic and knowledgeable answers for the questions in the following way:

\begin{itemize}
\item \textbf{Q1: How to capture the graph hierarchical structures in an interpretable way?}\\
The CommPOOL considers the communities as the basic graph hierarchical structure. In the community pooling operation, we adopt PAM to group graph nodes into different communities based on the distances among their features. Such a clustering-based community capture method is transparent enough to be interpreted. The community capture ability of our pooling operation has been shown in the previous \textbf{Community Evaluation on Simulation Data} section.     
\item \textbf{Q2: How to scale down the graph while preserving the structures via an interpretable process?} \\
To scale down the graph, we choose the community medoid node as the representation of the whole community, which can be understood like the centroid can be used to represent the whole mass. Meanwhile, without loss of necessary graph structure information, an interpretable structure preservation process is introduced during downscaling the graph. The community pooling achieves the preservation via gathering the nodal and structure information of the community member nodes as the features of the community medoid node. 
\item \textbf{Q3: What do we obtain after the pooling operation?} \\
From the graph topology view, the community pooling generates community-based sub-graphs of the original graph since the pooling operation does not generate any new graph nodes and edges. Each node in the sub-graph contains the corresponding community information.  
\end{itemize}

\subsection{Community Effect on Graph Classification}
We design a further experiment named \textbf{\textsl{semi-random pooling}} to show that a solid community preservation is important to the graph classification. Instead of randomly partitioning  the graph into multiple communities, we only randomly select the community center nodes. After determining the community center nodes, we assign each other node to the closest community based on the similarity of node features. Such a semi-random partition method can generate a few node cliques in graphs. These cliques, though are not the optimal communities, can still maintain the hierarchical information to some degree. We replace the PAM clustering by the semi-random partition in the pooling operation.  \textbf{Table~\ref{semi-random}} indicates that the community pooling has significant improvements in the graph classification tasks comparing to the semi-random pooling, which demonstrates that the success of community capture and preservation is crucial to the graph classification. To visualize, we select two simulation graphs to show (1) a positive example of community structure captured by the CommPOOL (\textbf{Figure~\ref{community}b}); and (2) a negative example of community structure captured by the CommPOOL, which eventually leads to the graph's misclassification (\textbf{Figure~\ref{community}c}). In addition, the performance of semi-random pooling does not decrease a lot comparing with the community pooling, which is beyond our expectations in a way. A reasonable explanation is that although unable to preserve the optimal community structure, the semi-random pooling method can still capture some degree of graph hierarchical structure, which again justifies that the significance of the community structure in the graph.
\vspace{-0.4em}
\begin{table}[h]
\large
    \centering
    \caption{Graph Classification Accuracy of Semi-random Pooling vs. Community-based Pooling}
    \label{tab:my_label}
    \begin{tabular}{c|c|c}
         \hline
         \textbf{Dataset} & \textbf{semi-rand.} & \textbf{community-based} \\
         \hline
         PROTEINS & $64.90\pm2.45$ & $74.74\pm0.06$ \\
         BZR & $81.50\pm2.82$ & $86.00\pm1.23$ \\
         Synthie & $59.00\pm5.89$ & $66.50\pm0.38$ \\
         Simulation & $70.34\pm1.26$ & $80.14\pm2.15$ \\
         \hline
    \end{tabular}
    \label{semi-random}
\end{table}
\vspace{-0.4em}
\section{Conclusion}
In this paper, we propose CommPOOL, a new interpretable hierarchical graph pooling framework. CommPOOL is designed for being able to capture and preserve the inherent hierarchical community structures in graphs during the graph representation learning and scaling-down process. Moreover, CommPOOL is a general graph representation learning framework that can facilitate various graph-level tasks. Experiments on both real-world graph datasets from different domains and synthetic graph data have shown that CommPOOL outperforms the state-of-the-art methods in graph representation learning for the graph classification task. In future work, we will explore leveraging CommPOOL for other graph-level tasks, such as graph regression.  

\bibliography{main.bib}

\begin{thebibliography}{56}
\providecommand{\natexlab}[1]{#1}
\providecommand{\url}[1]{\texttt{#1}}
\providecommand{\urlprefix}{URL }
\expandafter\ifx\csname urlstyle\endcsname\relax
  \providecommand{\doi}[1]{doi:\discretionary{}{}{}#1}\else
  \providecommand{\doi}{doi:\discretionary{}{}{}\begingroup
  \urlstyle{rm}\Url}\fi

\bibitem[{Bianchi, Grattarola, and Alippi(2020)}]{Bianchi2020SpectralCW}
Bianchi, F.~M.; Grattarola, D.; and Alippi, C. 2020.
\newblock Spectral Clustering with Graph Neural Networks for Graph Pooling.
\newblock \emph{arXiv: Learning} .

\bibitem[{Borgwardt et~al.(2005)Borgwardt, Ong, Sch{\"o}nauer, Vishwanathan,
  Smola, and Kriegel}]{borgwardt2005protein}
Borgwardt, K.~M.; Ong, C.~S.; Sch{\"o}nauer, S.; Vishwanathan, S.; Smola,
  A.~J.; and Kriegel, H.-P. 2005.
\newblock Protein function prediction via graph kernels.
\newblock \emph{Bioinformatics} 21(suppl\_1): i47--i56.

\bibitem[{Brandes, Gaertler, and Wagner(2003)}]{brandes2003experiments}
Brandes, U.; Gaertler, M.; and Wagner, D. 2003.
\newblock Experiments on graph clustering algorithms.
\newblock In \emph{European Symposium on Algorithms}, 568--579. Springer.

\bibitem[{Bruna et~al.(2013)Bruna, Zaremba, Szlam, and
  LeCun}]{bruna2013spectral}
Bruna, J.; Zaremba, W.; Szlam, A.; and LeCun, Y. 2013.
\newblock Spectral networks and locally connected networks on graphs.
\newblock \emph{arXiv preprint arXiv:1312.6203} .

\bibitem[{Chen, Ma, and Xiao(2018)}]{chen2018fastgcn}
Chen, J.; Ma, T.; and Xiao, C. 2018.
\newblock Fastgcn: fast learning with graph convolutional networks via
  importance sampling.
\newblock \emph{arXiv preprint arXiv:1801.10247} .

\bibitem[{Dai, Dai, and Song(2016)}]{dai2016discriminative}
Dai, H.; Dai, B.; and Song, L. 2016.
\newblock Discriminative embeddings of latent variable models for structured
  data.
\newblock In \emph{International conference on machine learning}, 2702--2711.

\bibitem[{Dai et~al.(2018)Dai, Li, Tian, Huang, Wang, Zhu, and
  Song}]{dai2018adversarial}
Dai, H.; Li, H.; Tian, T.; Huang, X.; Wang, L.; Zhu, J.; and Song, L. 2018.
\newblock Adversarial attack on graph structured data.
\newblock \emph{arXiv preprint arXiv:1806.02371} .

\bibitem[{Dhillon, Guan, and Kulis(2007)}]{dhillon2007weighted}
Dhillon, I.~S.; Guan, Y.; and Kulis, B. 2007.
\newblock Weighted graph cuts without eigenvectors a multilevel approach.
\newblock \emph{IEEE transactions on pattern analysis and machine intelligence}
  29(11): 1944--1957.

\bibitem[{Dobson and Doig(2003)}]{dobson2003distinguishing}
Dobson, P.~D.; and Doig, A.~J. 2003.
\newblock Distinguishing enzyme structures from non-enzymes without alignments.
\newblock \emph{Journal of molecular biology} 330(4): 771--783.

\bibitem[{Duvenaud et~al.(2015)Duvenaud, Maclaurin, Iparraguirre, Bombarell,
  Hirzel, Aspuru-Guzik, and Adams}]{duvenaud2015convolutional}
Duvenaud, D.~K.; Maclaurin, D.; Iparraguirre, J.; Bombarell, R.; Hirzel, T.;
  Aspuru-Guzik, A.; and Adams, R.~P. 2015.
\newblock Convolutional networks on graphs for learning molecular fingerprints.
\newblock In \emph{Advances in neural information processing systems},
  2224--2232.

\bibitem[{Fey and Lenssen(2019)}]{fey2019fast}
Fey, M.; and Lenssen, J.~E. 2019.
\newblock Fast graph representation learning with PyTorch Geometric.
\newblock \emph{arXiv preprint arXiv:1903.02428} .

\bibitem[{Fortunato(2010)}]{fortunato2010community}
Fortunato, S. 2010.
\newblock Community detection in graphs.
\newblock \emph{Physics reports} 486(3-5): 75--174.

\bibitem[{Gao and Ji(2019)}]{gao2019graph}
Gao, H.; and Ji, S. 2019.
\newblock Graph u-nets.
\newblock \emph{arXiv preprint arXiv:1905.05178} .

\bibitem[{Gilmer et~al.(2017)Gilmer, Schoenholz, Riley, Vinyals, and
  Dahl}]{gilmer2017neural}
Gilmer, J.; Schoenholz, S.~S.; Riley, P.~F.; Vinyals, O.; and Dahl, G.~E. 2017.
\newblock Neural message passing for quantum chemistry.
\newblock \emph{arXiv preprint arXiv:1704.01212} .

\bibitem[{Girvan and Newman(2002)}]{girvan2002community}
Girvan, M.; and Newman, M.~E. 2002.
\newblock Community structure in social and biological networks.
\newblock \emph{Proceedings of the national academy of sciences} 99(12):
  7821--7826.

\bibitem[{Hamilton, Ying, and Leskovec(2017)}]{hamilton2017inductive}
Hamilton, W.; Ying, Z.; and Leskovec, J. 2017.
\newblock Inductive representation learning on large graphs.
\newblock In \emph{Advances in neural information processing systems},
  1024--1034.

\bibitem[{Henaff, Bruna, and LeCun(2015)}]{henaff2015deep}
Henaff, M.; Bruna, J.; and LeCun, Y. 2015.
\newblock Deep convolutional networks on graph-structured data. arXiv 2015.
\newblock \emph{arXiv preprint arXiv:1506.05163} .

\bibitem[{Hou et~al.(2019)Hou, Zhang, Cheng, Ma, Ma, Chen, and
  Yang}]{hou2019measuring}
Hou, Y.; Zhang, J.; Cheng, J.; Ma, K.; Ma, R.~T.; Chen, H.; and Yang, M.-C.
  2019.
\newblock Measuring and improving the use of graph information in graph neural
  networks.
\newblock In \emph{International Conference on Learning Representations}.

\bibitem[{Huang et~al.(2018)Huang, Zhang, Rong, and Huang}]{huang2018adaptive}
Huang, W.; Zhang, T.; Rong, Y.; and Huang, J. 2018.
\newblock Adaptive sampling towards fast graph representation learning.
\newblock In \emph{Advances in neural information processing systems},
  4558--4567.

\bibitem[{Kaufmann(1987)}]{kaufmann1987clustering}
Kaufmann, L. 1987.
\newblock Clustering by means of medoids.
\newblock In \emph{Proc. Statistical Data Analysis Based on the L1 Norm
  Conference, Neuchatel, 1987}, 405--416.

\bibitem[{Kefato and Girdzijauskas(2020)}]{kefato2020graph}
Kefato, Z.~T.; and Girdzijauskas, S. 2020.
\newblock Graph neighborhood attentive pooling.
\newblock \emph{arXiv preprint arXiv:2001.10394} .

\bibitem[{Kersting et~al.(2016)Kersting, Kriege, Morris, Mutzel, and
  Neumann}]{KKMMN2016}
Kersting, K.; Kriege, N.~M.; Morris, C.; Mutzel, P.; and Neumann, M. 2016.
\newblock Benchmark Data Sets for Graph Kernels.
\newblock \urlprefix\url{http://graphkernels.cs.tu-dortmund.de}.

\bibitem[{Kipf and Welling(2016{\natexlab{a}})}]{kipf2016semi}
Kipf, T.~N.; and Welling, M. 2016{\natexlab{a}}.
\newblock Semi-supervised classification with graph convolutional networks.
\newblock \emph{arXiv preprint arXiv:1609.02907} .

\bibitem[{Kipf and Welling(2016{\natexlab{b}})}]{kipf2016variational}
Kipf, T.~N.; and Welling, M. 2016{\natexlab{b}}.
\newblock Variational graph auto-encoders.
\newblock \emph{arXiv preprint arXiv:1611.07308} .

\bibitem[{Kong and Yu(2014)}]{kong2014brain}
Kong, X.; and Yu, P.~S. 2014.
\newblock Brain network analysis: a data mining perspective.
\newblock \emph{ACM SIGKDD Explorations Newsletter} 15(2): 30--38.

\bibitem[{Lee et~al.(2019)Lee, Shen, Hou, and Hsu}]{lee2019high}
Lee, C.-F.; Shen, J.-J.; Hou, K.-L.; and Hsu, F.-W. 2019.
\newblock A High-performance Computing Method for Photographic Mosaics upon the
  Hadoop Framework.
\newblock \emph{Journal of Internet Technology} 20(5): 1343--1358.

\bibitem[{Lee, Lee, and Kang(2019)}]{lee2019self}
Lee, J.; Lee, I.; and Kang, J. 2019.
\newblock Self-attention graph pooling.
\newblock \emph{arXiv preprint arXiv:1904.08082} .

\bibitem[{Levie et~al.(2018)Levie, Monti, Bresson, and
  Bronstein}]{levie2018cayleynets}
Levie, R.; Monti, F.; Bresson, X.; and Bronstein, M.~M. 2018.
\newblock Cayleynets: Graph convolutional neural networks with complex rational
  spectral filters.
\newblock \emph{IEEE Transactions on Signal Processing} 67(1): 97--109.

\bibitem[{Li et~al.(2015)Li, Tarlow, Brockschmidt, and Zemel}]{li2015gated}
Li, Y.; Tarlow, D.; Brockschmidt, M.; and Zemel, R. 2015.
\newblock Gated graph sequence neural networks.
\newblock \emph{arXiv preprint arXiv:1511.05493} .

\bibitem[{Liu et~al.(2019)Liu, Ma, Jiang, Lu, Philip, and
  Ragin}]{liu2019community}
Liu, J.; Ma, G.; Jiang, F.; Lu, C.-T.; Philip, S.~Y.; and Ragin, A.~B. 2019.
\newblock Community-preserving graph convolutions for structural and functional
  joint embedding of brain networks.
\newblock In \emph{2019 IEEE International Conference on Big Data (Big Data)},
  1163--1168. IEEE.

\bibitem[{Long et~al.(2019)Long, Wang, Du, Song, Jin, and
  Lin}]{long2019hierarchical}
Long, Q.; Wang, Y.; Du, L.; Song, G.; Jin, Y.; and Lin, W. 2019.
\newblock Hierarchical community structure preserving network embedding: A
  subspace approach.
\newblock In \emph{Proceedings of the 28th ACM International Conference on
  Information and Knowledge Management}, 409--418.

\bibitem[{Ma et~al.(2019{\natexlab{a}})Ma, Ahmed, Willke, Sengupta, Cole,
  Turk-Browne, and Yu}]{ma2019deep}
Ma, G.; Ahmed, N.~K.; Willke, T.~L.; Sengupta, D.; Cole, M.~W.; Turk-Browne,
  N.~B.; and Yu, P.~S. 2019{\natexlab{a}}.
\newblock Deep graph similarity learning for brain data analysis.
\newblock In \emph{Proceedings of the 28th ACM International Conference on
  Information and Knowledge Management}, 2743--2751.

\bibitem[{Ma et~al.(2019{\natexlab{b}})Ma, Ahmed, Willke, and
  Yu}]{ma2019survey}
Ma, G.; Ahmed, N.~K.; Willke, T.~L.; and Yu, P.~S. 2019{\natexlab{b}}.
\newblock Deep Graph Similarity Learning: A Survey.
\newblock \emph{arXiv preprint arXiv:1912.11615} .

\bibitem[{Ma et~al.(2019{\natexlab{c}})Ma, Wang, Aggarwal, and
  Tang}]{ma2019graph}
Ma, Y.; Wang, S.; Aggarwal, C.~C.; and Tang, J. 2019{\natexlab{c}}.
\newblock Graph convolutional networks with eigenpooling.
\newblock In \emph{Proceedings of the 25th ACM SIGKDD International Conference
  on Knowledge Discovery \& Data Mining}, 723--731.

\bibitem[{Meunier et~al.(2009)Meunier, Lambiotte, Fornito, Ersche, and
  Bullmore}]{meunier2009hierarchical}
Meunier, D.; Lambiotte, R.; Fornito, A.; Ersche, K.; and Bullmore, E.~T. 2009.
\newblock Hierarchical modularity in human brain functional networks.
\newblock \emph{Frontiers in neuroinformatics} 3: 37.

\bibitem[{Miller(2019)}]{miller2019explanation}
Miller, T. 2019.
\newblock Explanation in artificial intelligence: Insights from the social
  sciences.
\newblock \emph{Artificial Intelligence} 267: 1--38.

\bibitem[{Molnar(2020)}]{molnar2020interpretable}
Molnar, C. 2020.
\newblock \emph{Interpretable Machine Learning}.
\newblock Lulu. com.

\bibitem[{Morris et~al.(2016)Morris, Kriege, Kersting, and
  Mutzel}]{morris2016faster}
Morris, C.; Kriege, N.~M.; Kersting, K.; and Mutzel, P. 2016.
\newblock Faster kernels for graphs with continuous attributes via hashing.
\newblock In \emph{2016 IEEE 16th International Conference on Data Mining
  (ICDM)}, 1095--1100. IEEE.

\bibitem[{Orsini, Frasconi, and De~Raedt(2015)}]{orsini2015graph}
Orsini, F.; Frasconi, P.; and De~Raedt, L. 2015.
\newblock Graph invariant kernels.
\newblock In \emph{Proceedings of the twenty-fourth international joint
  conference on artificial intelligence}, volume 2015, 3756--3762. IJCAI-INT
  JOINT CONF ARTIF INTELL.

\bibitem[{Paszke et~al.(2017)Paszke, Gross, Chintala, Chanan, Yang, DeVito,
  Lin, Desmaison, Antiga, and Lerer}]{paszke2017automatic}
Paszke, A.; Gross, S.; Chintala, S.; Chanan, G.; Yang, E.; DeVito, Z.; Lin, Z.;
  Desmaison, A.; Antiga, L.; and Lerer, A. 2017.
\newblock Automatic differentiation in pytorch .

\bibitem[{Riesen and Bunke(2008)}]{riesen2008iam}
Riesen, K.; and Bunke, H. 2008.
\newblock IAM graph database repository for graph based pattern recognition and
  machine learning.
\newblock In \emph{Joint IAPR International Workshops on Statistical Techniques
  in Pattern Recognition (SPR) and Structural and Syntactic Pattern Recognition
  (SSPR)}, 287--297. Springer.

\bibitem[{Spirin and Mirny(2003)}]{spirin2003protein}
Spirin, V.; and Mirny, L.~A. 2003.
\newblock Protein complexes and functional modules in molecular networks.
\newblock \emph{Proceedings of the national Academy of sciences} 100(21):
  12123--12128.

\bibitem[{Strehl and Ghosh(2002)}]{strehl2002cluster}
Strehl, A.; and Ghosh, J. 2002.
\newblock Cluster ensembles---a knowledge reuse framework for combining
  multiple partitions.
\newblock \emph{Journal of machine learning research} 3(Dec): 583--617.

\bibitem[{Sutherland, O'brien, and Weaver(2003)}]{sutherland2003spline}
Sutherland, J.~J.; O'brien, L.~A.; and Weaver, D.~F. 2003.
\newblock Spline-fitting with a genetic algorithm: A method for developing
  classification structure- activity relationships.
\newblock \emph{Journal of chemical information and computer sciences} 43(6):
  1906--1915.

\bibitem[{Tang et~al.(2020)Tang, Ma, Chen, Guo, Wang, Zeng, and
  Zhan}]{tang2020adversarial}
Tang, H.; Ma, G.; Chen, Y.; Guo, L.; Wang, W.; Zeng, B.; and Zhan, L. 2020.
\newblock Adversarial Attack on Hierarchical Graph Pooling Neural Networks.
\newblock \emph{arXiv preprint arXiv:2005.11560} .

\bibitem[{Tran, Navarin, and Sperduti(2018)}]{tran2018filter}
Tran, D.~V.; Navarin, N.; and Sperduti, A. 2018.
\newblock On filter size in graph convolutional networks.
\newblock In \emph{2018 IEEE Symposium Series on Computational Intelligence
  (SSCI)}, 1534--1541. IEEE.

\bibitem[{Veli{\v{c}}kovi{\'c} et~al.(2017)Veli{\v{c}}kovi{\'c}, Cucurull,
  Casanova, Romero, Lio, and Bengio}]{velivckovic2017graph}
Veli{\v{c}}kovi{\'c}, P.; Cucurull, G.; Casanova, A.; Romero, A.; Lio, P.; and
  Bengio, Y. 2017.
\newblock Graph attention networks.
\newblock \emph{arXiv preprint arXiv:1710.10903} .

\bibitem[{Vinyals, Bengio, and Kudlur(2015)}]{vinyals2015order}
Vinyals, O.; Bengio, S.; and Kudlur, M. 2015.
\newblock Order matters: Sequence to sequence for sets.
\newblock \emph{arXiv preprint arXiv:1511.06391} .

\bibitem[{Wu et~al.(2020)Wu, Pan, Chen, Long, Zhang, and
  Yu}]{wu2020comprehensive}
Wu, Z.; Pan, S.; Chen, F.; Long, G.; Zhang, C.; and Yu, S.~P. 2020.
\newblock A comprehensive survey on graph neural networks.
\newblock \emph{IEEE Transactions on Neural Networks and Learning Systems} .

\bibitem[{Ying et~al.(2019)Ying, Bourgeois, You, Zitnik, and
  Leskovec}]{ying2019gnnexplainer}
Ying, Z.; Bourgeois, D.; You, J.; Zitnik, M.; and Leskovec, J. 2019.
\newblock Gnnexplainer: Generating explanations for graph neural networks.
\newblock In \emph{Advances in neural information processing systems},
  9244--9255.

\bibitem[{Ying et~al.(2018)Ying, You, Morris, Ren, Hamilton, and
  Leskovec}]{ying2018hierarchical}
Ying, Z.; You, J.; Morris, C.; Ren, X.; Hamilton, W.; and Leskovec, J. 2018.
\newblock Hierarchical graph representation learning with differentiable
  pooling.
\newblock In \emph{Advances in neural information processing systems},
  4800--4810.

\bibitem[{Yuan et~al.(2020)Yuan, Tang, Hu, and Ji}]{yuan2020xgnn}
Yuan, H.; Tang, J.; Hu, X.; and Ji, S. 2020.
\newblock XGNN: Towards Model-Level Explanations of Graph Neural Networks.
\newblock \emph{arXiv preprint arXiv:2006.02587} .

\bibitem[{Zhang et~al.(2018)Zhang, Cui, Neumann, and Chen}]{zhang2018end}
Zhang, M.; Cui, Z.; Neumann, M.; and Chen, Y. 2018.
\newblock An end-to-end deep learning architecture for graph classification.
\newblock In \emph{Thirty-Second AAAI Conference on Artificial Intelligence}.

\bibitem[{Zhang et~al.(2019)Zhang, Bu, Ester, Zhang, Yao, Yu, and
  Wang}]{zhang2019hierarchical}
Zhang, Z.; Bu, J.; Ester, M.; Zhang, J.; Yao, C.; Yu, Z.; and Wang, C. 2019.
\newblock Hierarchical graph pooling with structure learning.
\newblock \emph{arXiv preprint arXiv:1911.05954} .

\bibitem[{Z{\"u}gner, Akbarnejad, and
  G{\"u}nnemann(2018)}]{zugner2018adversarial}
Z{\"u}gner, D.; Akbarnejad, A.; and G{\"u}nnemann, S. 2018.
\newblock Adversarial attacks on neural networks for graph data.
\newblock In \emph{Proceedings of the 24th ACM SIGKDD International Conference
  on Knowledge Discovery \& Data Mining}, 2847--2856.

\bibitem[{Z{\"u}gner and G{\"u}nnemann(2020)}]{zugner2020certifiable}
Z{\"u}gner, D.; and G{\"u}nnemann, S. 2020.
\newblock Certifiable Robustness of Graph Convolutional Networks under
  Structure Perturbations.
\newblock In \emph{Proceedings of the 26th ACM SIGKDD International Conference
  on Knowledge Discovery \& Data Mining}, 1656--1665.

\end{thebibliography}
\end{document}